\documentclass[sigconf]{acmart}

\usepackage{color}

\newif\ifworkinprogress
\workinprogresstrue

\ifworkinprogress
  \newcommand{\mq}[1]{\textcolor{blue}{\textbf{[Massimo] #1}}}
  \newcommand{\ak}[1]{\textcolor{magenta}{\textbf{[Alex] #1}}}
  \newcommand{\bh}[1]{\textcolor{yellow}{\textbf{[Bal\'{a}zs] #1}}}
  \newcommand{\pc}[1]{\textcolor{cyan}{\textbf{[Paolo] #1}}}
\else
  \newcommand{\mq}[1]{}
  \newcommand{\ak}[1]{}
  \newcommand{\bh}[1]{}
  \newcommand{\pc}[1]{}
\fi

\usepackage{csquotes}
\usepackage[T1]{fontenc}
\usepackage{epsfig}
\usepackage{color}

\usepackage{balance}  
\usepackage{algorithmic}
\usepackage{algorithm}
\usepackage{multirow}
\usepackage{graphicx}
\usepackage{xspace}
\usepackage{url}
\usepackage{hyperref}
\usepackage{booktabs}
\usepackage{slantsc}
\usepackage{array}
\newcommand{\normaltilde}{\raise.17ex\hbox{$\scriptstyle\sim$}}
\newcommand*\rot{\rotatebox{90}}
\newcolumntype{P}[1]{>{\centering\arraybackslash}p{#1}}

\newcommand{\RNNt}{Recurrent Neural Network}
\newcommand{\RNN}{RNN}
\newcommand{\RNNConcat}{\RNN{} Concat}

\newcommand{\HRNNt}{Hierarchical RNN}
\newcommand{\HRNN}{\textit {HRNN}}
\newcommand{\HRNNInit}{\textit {HRNN Init}}
\newcommand{\HRNNAll}{\textit {HRNN All}}

\newcommand{\GRUt}{Gated Recurrent Unit}
\newcommand{\GRU}{GRU}
\newcommand{\GRUses}{$GRU_{ses}$}

\newcommand{\GRUusr}{$GRU_{usr}$}

\newcommand{\PPOP}{PPOP}
\newcommand{\ItemKNN}{Item-KNN}

\newcommand{\ita}[1]{\textsl{#1}}

\newcommand{\iu}[1]{{\slshape{\underline{#1}}}}

\newcommand{\iub}[1]{{\slshape{\underline{\textbf{#1}}}}}

\newcommand{\IGNORE}[1]{}

\newcommand{\commentedtext}[1]{}

\graphicspath{{./figures/}}

\copyrightyear{2017} 
\acmYear{2017} 
\setcopyright{acmcopyright}
\acmConference{RecSys'17}{}{August 27--31, 2017, Como, Italy.}
\acmPrice{15.00}
\acmDOI{http://dx.doi.org/10.1145/3109859.3109896}
\acmISBN{978-1-4503-4652-8/17/08}
\begin{document}

\title[Personalizing Session-based Recommendations with Hierarchical RNNs]{Personalizing Session-based Recommendations with Hierarchical Recurrent Neural Networks}

\author{Massimo Quadrana}
\affiliation{Politecnico di Milano, Milan, Italy}
\email{massimo.quadrana@polimi.it}
\author{Alexandros Karatzoglou} 
\affiliation{Telefonica Research, Barcelona, Spain}
\email{alexk@tid.es}
\author{Bal\'{a}zs Hidasi}
\affiliation{Gravity R\&D, Budapest, Hungary}
\email{balazs.hidasi@gravityrd.com}
\author{Paolo Cremonesi}
\affiliation{Politecnico di Milano, Milan, Italy}
\email{paolo.cremonesi@polimi.it}
\renewcommand{\shortauthors}{M. Quadrana et al.}

\begin{abstract}
Session-based recommendations are highly relevant in many modern on-line services (e.g. e-commerce, video streaming) and recommendation settings. Recently, Recurrent Neural Networks have been shown to perform very well in session-based settings. While in many session-based recommendation domains user identifiers are hard to come by, there are also domains in which user profiles are readily available. We propose a seamless way to personalize RNN models with cross-session information transfer and devise a Hierarchical RNN model that relays end evolves latent hidden states of the RNNs across user sessions. 
Results on two industry datasets show large improvements over the session-only RNNs.
\end{abstract}

%
%


\maketitle

\section{Introduction}
\label{sec:introduction}
In many online systems where recommendations are applied, interactions between a user and the system are organized into sessions. 
A session is a group of interactions that take place within a given time frame. 
Sessions from a user can occur on the same day, or over several days, weeks, or months. 
A session usually has a goal, such as finding a good restaurant in a city, or listening to music of a certain style or mood.

Providing recommendations in these domains poses unique challenges that until recently have been mainly tackled by applying conventional recommender algorithms~\cite{item-oriented-2013} on either the last interaction or the last session (session-based recommenders). 
Recurrent Neural Networks (RNNs) have been recently used for the purpose of session-based recommendations~\cite{hidasi16session} outperforming item-based methods by 15\% to 30\% in terms of ranking metrics. 
In \textit{session-based recommenders}, recommendations are provided based solely on the interactions in the current user session, as user are assumed to be anonymous. 
But in many of these systems there are cases where a user might be logged-in (e.g. music streaming services) or some form of user identifier might be present (cookie or other identifier). In these cases it is reasonable to assume that the user behavior in past sessions might provide valuable information for providing recommendations in the next session. 

A simple way of incorporating past user session information in session-based algorithm would be to simply concatenate past and current user sessions. 
While this seems like a reasonable approach, we will see in the experimental section that this does not yield the best results. 

In this work we describe a novel algorithm based on RNNs that can deal with both cases: (i) \textit{session-aware recommenders}, when user identifiers are present and propagate information from the previous user session to the next, thus improving the recommendation accuracy, and (ii) \textit{session-based recommenders}, when there are no past sessions (i.e., no user identifiers). 
The algorithm is based on a Hierarchical RNN where the hidden state of a lower-level RNN at the end of one user session is passed as an input to a higher-level RNN which aims at predicting a good initialization (i.e., a good context vector) for the hidden state of the lower RNN for the next session of the user.

We evaluate the Hierarchical RNNs on two datasets from industry comparing them to the plain session-based RNN and to item-based collaborative filtering. Hierarchical RNNs outperform both alternatives by a healthy margin.

\section{Related work}
\label{sec:related}
\textbf{Session-based recommendations.}
Classical CF methods (e.g.\ matrix factorization) break down in the session-based setting when no user profile can be constructed from past user behavior. A natural solution to this problem is the item-to-item recommendation approach \cite{sarwar2001item,linden2003amazon}. In this setting an item-to-item similarity matrix is precomputed from the available session data, items that are often clicked together in sessions are deemed to be similar. These similarities are then used to create recommendations. While simple, this
method has been proven to be effective and is widely employed. Though, these methods only take into account the last click of the user, in effect ignoring the information of the previous clicks.

\textbf{Recurrent Neural Models.} RNNs are the deep models of choice when dealing with sequential data \cite{lipton2015critical}. RNNs have been used in image and video captioning, time series prediction, natural language processing
and much more. Long Short-Term Memory (LSTM) \cite{hochreiter1997long} networks are a type of RNNs that have been shown to work particularly well, it includes additional gates that regulate when and how much to take the input into account and when to reset the hidden state. This helps with the vanishing gradient problem that often plagues the standard RNN models. Slightly simplified version of LSTM -- that still maintains all their properties -- are Gated Recurrent Units (GRUs) \cite{cho2014properties} which we use in this work.


RNNs were first used to model session data in \cite{hidasi16session}. The recurrent neural network is trained with a ranking loss on a one-hot representation of the session (clicked) item-IDs.  The RNN is then used to provide recommendations after each click for new sessions. This work only focused on the clicked item-IDs in the current session while here we aim at modeling the user behavior across sessions as well. RNNs were also used to jointly model the content or features of items together with click-sequence interactions \cite{hidasi16feature}. By including item features extracted for example, from the thumbnail image of videos or the textual description of a product, the so-called parallel-RNN model provided superior recommendation quality wrt.\ `plain' RNNs.
In \cite{improved_rnn}, authors proposed data augmentation techniques to improve the performance of the RNN for session-based recommendations, these techniques have though the side effect of  increasing training times as a single session is split into several sub-sessions for training. 
RNNs have also been used in more standard user-item collaborative filtering settings where the aim is to model the evolution of the user and items factors~\cite{Wu:2017}~\cite{devooght2016collaborative} where the results are though less impressive, with the proposed methods barely outperforming standard matrix factorization methods. Finally a sequence to sequence model with a version of Hierarchical Recurrent Neural Networks was used for generative context-aware query suggestion in \cite{sordoni2015hierarchical}.

\section{Model}
\label{sec:model}
In this section we describe the proposed \HRNNt{} (\HRNN{} henceforth) model for personalized session-based recommendation. 

\subsection{Session-based Recurrent Neural Network} 
Our model is based on the session-based \RNNt{} (\RNN{} henceforth) model presented in \cite{hidasi16session}. \RNN{} is based on a single \GRUt{} (\GRU{}) layer that models the interactions of the user within a session. The \RNN{} takes as input the current item ID in the session and outputs a score for each item representing the likelihood of being the next item in the session.

Formally, for each session $S_m = \{i_{m,1}, i_{m,2},...,i_{m,N_m}\}$, \RNN{} computes the following session-level representation
\begin{equation}
s_{m,n} = GRU_{ses}\left(i_{m,n}, s_{m,n-1}\right),\text{ } n=1,...,N_m-1
\end{equation}
where \GRUses{} is the \textit{session-level} \GRU{} and $s_{m,n}$ its hidden state at step $n$, being $s_{m,0} = 0$ (the null vector), and $i_{m,n}$ is the one-hot vector of the current item ID \footnote{To simplify the explanation we use a notation similar to \cite{sordoni2015hierarchical}.}.

The output of the \RNN{} is a score $\hat{r}_{m,n}$ for every item in the catalog indicating the likelihood of being the next item in the session (or, equivalently, its relevance for the next step in the session)
\begin{equation}\label{eq:prediction}
\hat{r}_{m,n} = g\left(s_{m,n}\right),\text{ } n=1,...,N_m-1
\end{equation}
where $g\left(\cdot\right)$ is a non-linear function like softmax or tanh depending on the loss function.
During training, scores are compared to a one-hot vector of the next item ID in the session to compute the loss. The network can be trained with several ranking loss functions such as cross-entropy, BPR \cite{rendle09bpr} and TOP1 \cite{hidasi16session}. In this work, the TOP1 loss always outperformed other ranking losses, so we consider only it in the rest of the paper. 
The \textbf{TOP1} loss is the regularized approximation of the
 relative rank of the relevant item. The relative rank of the relevant item is given by $\frac{1}{N_S}\cdot \sum_{j=1}^{N_S}{I\{\hat{r}_{s,j}>\hat{r}_{s,i}\}}$  where $r_{s,j}$ is the score of a sampled `irrelevant item'. $I\{\cdot\}$ is approximated with a sigmoid. To force the scores of negative examples (`irrelevant items') towards zero a regularization term is added to the loss. The final loss function is as follows: $L_s=\frac{1}{N_S}\cdot\sum_{j=1}^
{N_S}{\sigma\left(\hat{r}_{s,j}-\hat{r}_{s,i}\right)+\sigma\left(\hat{r}_{s,j}^2\right)}$

\RNN{} is trained efficiently with session-parallel mini-batches. At each training step, the input to \GRUses{} is the stacked one-hot representation of the current item ID of a batch of sessions. The session-parallel mechanism keeps the pointers to the current item of every session in the mini-batch and resets the hidden state of the \RNN{} when sessions end. To further reduce the computational complexity, 
the loss is computed over the current item IDs and a sample of negative  items. Specifically,
the current item ID of each session is used as positive item and the IDs of the remaining sessions in the mini-batch as negative items when computing the loss. This makes explicit negative item sampling unnecessary and enables popularity-based sampling. 
However, since user-identifiers are unknown in pure session-based scenarios, there are good chances that negative samples will be `contaminated' by positive items the user interacts with in other sessions.

\subsection{Personalized Session-based Hierarchical Recurrent Neural Network}
Our \HRNN{} model builds on top of \RNN{} by: (i) adding an additional \GRU{} layer to model information across user sessions and to track the evolution of the user interests over time; (ii) using a powerful user-parallel mini-batch mechanism for efficient training.

\begin{figure*}[t]
\centering
\includegraphics[width=\textwidth]{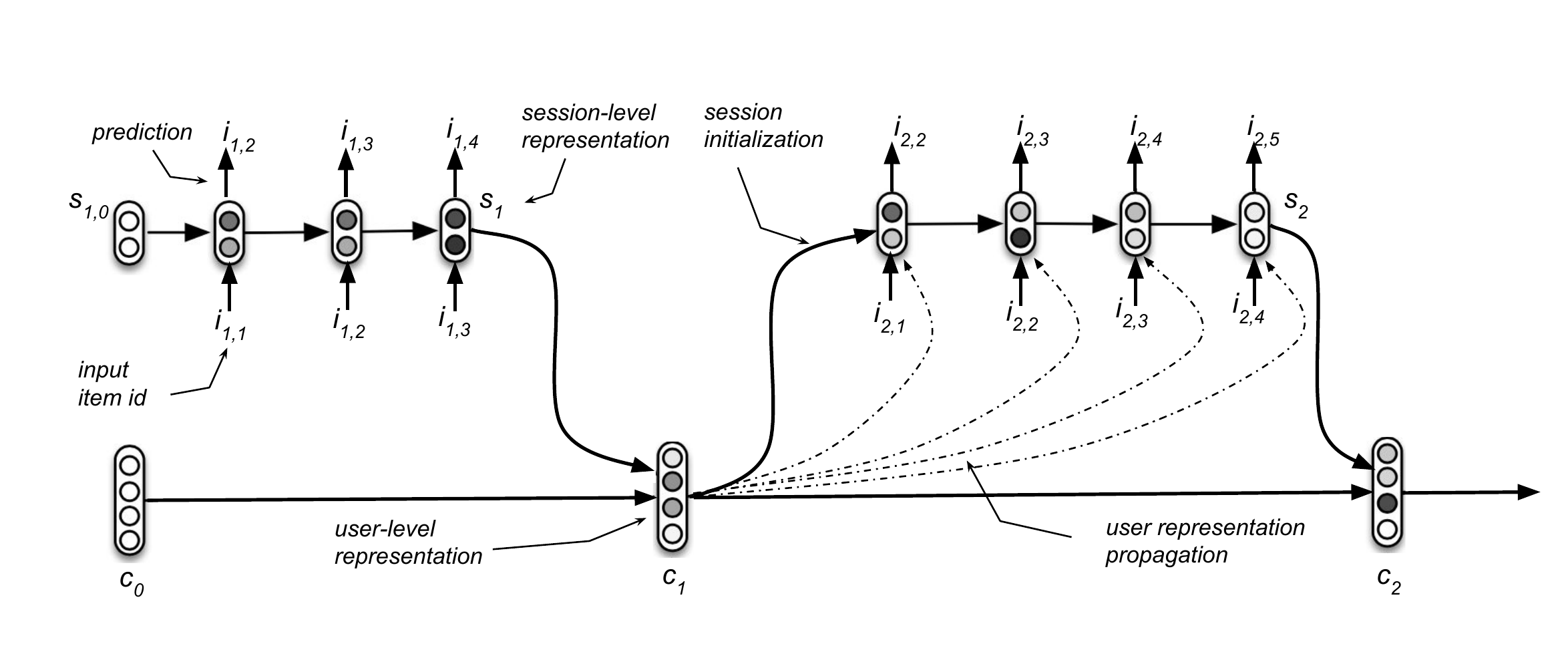}
\caption{Graphical representation of the proposed \HRNNt{} model for personalized session-based recommendation. The model is composed of an hierarchy of two \GRU{}s, the session-level \GRU{} (\GRUses{}) and the user-level \GRU{} (\GRUusr{}). The session-level \GRU{} models the user activity within sessions and generates recommendations. The user-level \GRU{} models the evolution of the user across sessions and provides personalization capabilities to the session-level \GRU{} by initializing its hidden state and, optionally, by propagating the user representation in input.}
\label{fig:hrnn_model}
\end{figure*}

\subsubsection{Architecture}
Beside the session-level \GRU{}, our \HRNN{} model adds one \textit{user-level} \GRU{} (\GRUusr{}) to model the user activity across sessions. 

\autoref{fig:hrnn_model} shows a graphical representation of \HRNN{}. At each time step, recommendations are generated by \GRUses{}, as in \RNN{}. However, when a session ends, the user representation is updated. When a new session starts, the hidden state of \GRUusr{} is used to initialize \GRUses{} and, optionally, propagated in input to \GRUses{}. 

Formally, for each user $u$ with sessions $C^u=\{S^u_1, S^u_2, ...,S^u_{M_u}\}$, the user-level \GRU{} takes as input the session-level representations $s^u_1, s^u_2, ..., s^u_{M_u}$, being $s^u_m = s^u_{m,N_m-1}$ the last hidden state of $GRU_{ses}$ of each user session $S^u_{m}$, and uses them to update the user-level representation $c^u_m$. Henceforth we drop the user superscript $u$ to unclutter notation. The user-level representation $c_m$ is updated as
\begin{equation}\label{eq:cti_update}
c_m = GRU_{usr}\left(s_m, c_{m-1}\right),\text{ } m=1,...,M_u
\end{equation}
where $c_0 = 0$ (the null vector). The input to the user-level \GRU{} is connected to the last hidden state of the session-level \GRU{}. In this way, the user-level \GRU{} can track the evolution of the user across sessions and, in turn, model the dynamics user interests seamlessly. Notice that the user-level representation is kept \textit{fixed} throughout the session and it is updated only when the session ends.

The user-level representation is then used to initialize the hidden state of the session-level \GRU{}. Given $c_m$, the initial hidden state $s_{m+1,0}$ of the session-level \GRU{} for the following session is set to
\begin{equation}\label{eq:ses_init}
s_{m+1,0} = \mathrm{tanh}\left(W_{init} c_m + b_{init}\right)
\end{equation}
where $W_{init}$ and $b_{init}$ are the initialization weights and biases respectively. In this way, the information relative to the preferences expressed by the user in the previous sessions is transferred to the session-level.
Session-level representations are then updated as follows
\begin{equation}
\label{eq:ses_update}
s_{m+1,n} = GRU_{ses}\left(i_{m+1,n}, s_{m+1,n-1}\left[, c_m\right] \right),\text{ } n=1,...,N_{m+1}-1
\end{equation}
where the square brackets indicate that $c_m$ can be \textit{optionally} propagated in input to the session-level \GRU{}.

The model is trained end-to-end using back-propagation \cite{rumelhart88backprop}. The weights of \GRUusr{} are updated only between sessions, i.e. when a session ends and when the forthcoming session starts.
However, when the user representation is propagated in input to \GRUses{}, the weights of \GRUusr{} are updated also within sessions even if $c_m$ is kept fixed. We also tried with propagating the user-level representation to the final prediction layer (i.e., by adding term $c_m$ in \autoref{eq:prediction}) but we always incurred into severe degradation of the performances, even wrt.\ simple session-based \RNN{}. We therefore discarded this setting from this discussion.

Note here that the \GRUusr{} does not simply pass on the hidden state of the previous user session to the next but also learns (during training) how user sessions evolve during time. We will see in the experimental section that this is crucial in achieving increased performance. In effect \GRUusr{} computes and evolves a user profile that is based on the previous user sessions, thus in effect personalizing the \GRUses{}. 
In the original \RNN{}, users who had clicked/interacted with the same sequence of items in a session would get the same recommendations; in \HRNN{} this is not anymore the case, recommendations will be influenced by the the users past sessions as well.   

In summary, we considered the following two different \HRNN{} settings, depending on whether the user representation $c_m$ is considered in \autoref{eq:ses_update}:
\begin{itemize}
\item \HRNNInit{}, in which $c_m$ is used only to initialize the representation of the next session.
\item \HRNNAll{}, in which $c_m$ is used for initialization and propagated in input at each step of the next session.
\end{itemize}
In \HRNNInit{}, the session-level \GRU{} can exploit the historical preferences along with the session-level dynamics of the user interest. \HRNNAll{} instead enforces the usage of the user representation at session-level at the expense of a slightly greater model complexity. As we will see, this can lead to substantially different results depending on the recommendation scenario.

\subsubsection{Learning}
\begin{figure}[t]
\centering
\includegraphics[width=\linewidth]{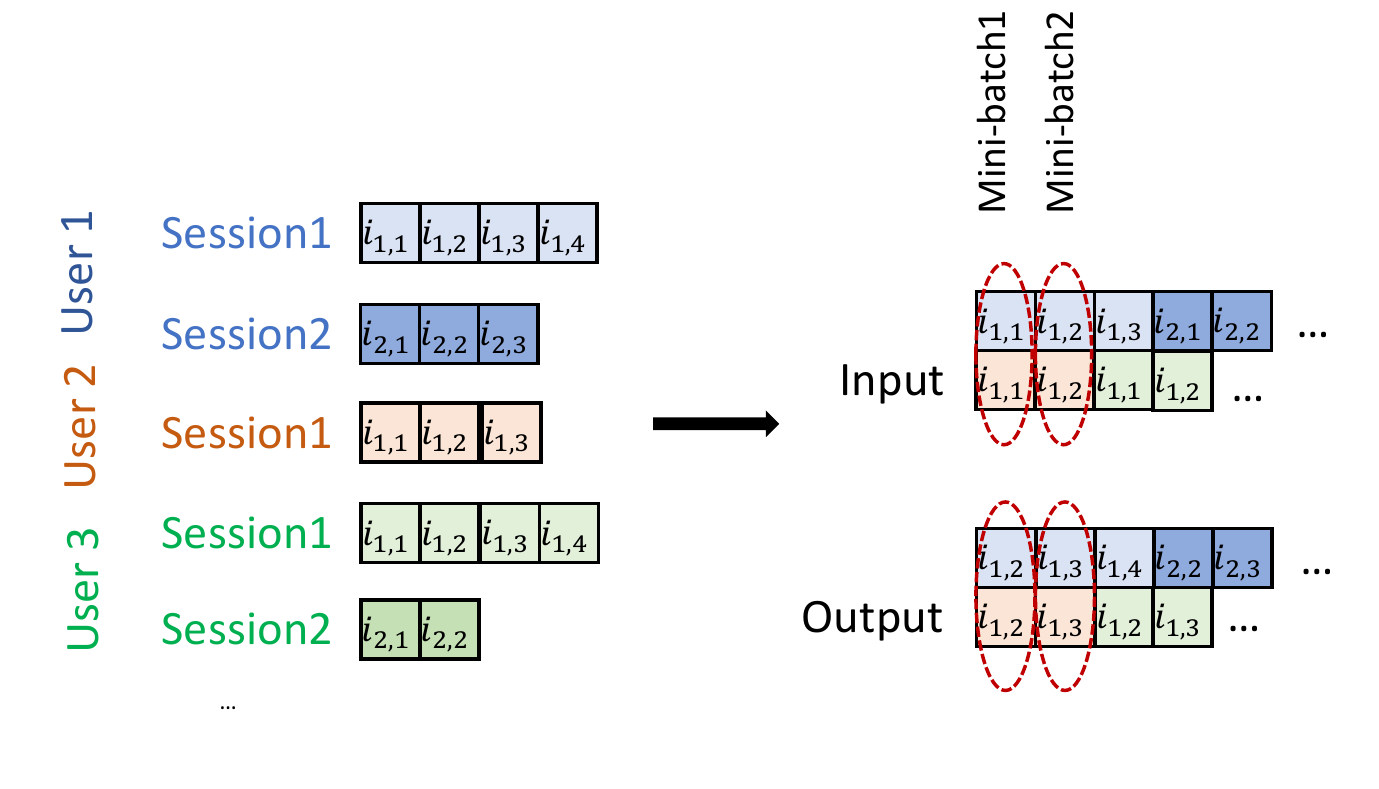}
\caption{User-parallel mini-batches for mini-batch size 2.}
\label{fig:minibatch}
\end{figure}

For the sake of efficiency in training, we have edited the session-parallel mini-batch mechanism described in \cite{hidasi16feature} to account for user identifiers during training (see \autoref{fig:minibatch}). We first group sessions by user and then sort session events within each group by time-stamp. We then order users at random. 
At the first iteration, the first item of the first session of the first $B$ users constitute the input to the \HRNN{}; the second item in each session constitute its output. The output is then used as input for the next iteration, and so on. When a session in the mini-batch ends, \autoref{eq:cti_update} is used to update the hidden state of \GRUusr{} and \autoref{eq:ses_init} to initialize the hidden state of \GRUses{} for the forthcoming session, if any. When a user has been processed completely, the hidden states of both \GRUusr{} and \GRUses{} are reset and the next user is put in its place in the mini-batch.

With user-parallel mini-batches we can train \HRNN{}s efficiently over users having different number of sessions and sessions of different length. Moreover, this mechanism allows to sample negative items in a user-independent fashion, hence reducing the chances of `contamination' of the negative samples with actual positive items. The sampling procedure is still popularity-based, since the likelihood for an item to appear in the mini-batch is proportional to its popularity. Both properties are known to be beneficial for pairwise learning with implicit user feedback \cite{rendle14improving}. 

\section{Experiments}
\label{sec:experiments}
In this section we describe our experimental setup and provide an in-depth discussion of the obtained results.
\subsection{Datasets}
We used two datasets for our experiments. The first is the XING \footnote{https://www.xing.com/en} Recsys Challenge 2016 dataset \cite{abel16challenge} that contains interactions on job postings for 770k users over a 80-days period. User interactions come with time-stamps and interaction type (click, bookmark, reply and delete). We named this dataset XING.
The second dataset is a proprietary dataset from a Youtube-like video-on-demand web site. 
The dataset tracks the videos watched by 13k users over a 2-months period. 
Viewing events lasting less than a fixed threshold are not tracked.
We named this dataset VIDEO.

We manually partitioned the interaction data into sessions by using a 30-minute idle threshold. For the XING dataset we discarded interactions having type `delete'. We also discarded repeated interactions of the same type within sessions to reduce noise (e.g. repeated clicks on the same job posting within a session).
We then preprocessed both datasets as follows. We removed items with support less than $20$ for XING and $10$ for VIDEO since items with low support are not optimal for modeling. We removed sessions having $<3$ interactions to filter too short and poorly informative sessions, and kept users having $\ge 5$ sessions to have sufficient cross-session information for proper modeling of returning users.

The test set is build with the last session of each user. The remaining sessions form the training set. We also filtered items in the test set that do not belong to the training set. This partitioning allows to run the evaluation over users having different amounts of historical sessions in their profiles, hence to measure the recommendation quality over users having different degrees of activity with the system (see Section~\ref{sec:eval_histlen} for an in-depth analysis).
We further partitioned the training set with the same procedure to tune the hyper-parameters of the algorithms. The characteristics of the datasets are summarized in \autoref{tab:datasets}.

\begin{table}[t]
\centering
\begin{tabular}{lcc}
\toprule
\textbf{Dataset} & \textbf{XING} & \textbf{VIDEO} \\
\midrule
Users & 11,479 & 13,717 \\
Items & 59,297 & 19,218 \\
Sessions & 89,591 & 133,165 \\
Events & 546,862 & 825,449 \\
Events per item$^\dagger{}$&6/9.3/14.4& 9/43.0/683.7\\
Events per session$^\dagger{}$ & 4/6.1/5.5 & 4/6.2/8.2 \\
Sessions per user$^\dagger{}$ & 6/7.8/4.8 & 7/9.8/7.8 \\
\midrule
Training - Events & 488,576 & 745,482 \\
Training - Sessions & 78,276 & 120,160 \\
Test - Events & 58,286 & 79,967 \\
Test - Sessions & 11,315 & 13,605\\
\bottomrule
\end{tabular}
\caption{Main properties of the datasets ($^\dagger{}$median/mean/std).}
\label{tab:datasets}
\end{table}
\subsection{Baselines and parameter tuning}
We compared our \HRNN{} model against several baselines, namely Personal Pop (\PPOP{}), \ItemKNN{}, \RNN{} and \RNNConcat{}: 
\begin{itemize}
\item 
Personal Pop (\PPOP{}) recommends the item with the largest number of interactions by the user.
\item 
\ItemKNN{} computes an item-to-item cosine similarity based on the co-occurrence of items within sessions.
\item 
\RNN{} adopts the same model described in \cite{hidasi16session}. 
This model uses the basic \GRU{} with a TOP1 loss function and session-parallel minibatching: sessions from the same user are fed to the \RNN{} independently from each other. 
\item \RNNConcat{} is the same as \RNN{}, but sessions from the same user are concatenated into a single session\footnote{We experimented with a variant of this baseline that adds a special `separator item' to enforce the separation between sessions; however, it did not show better performance in our experiments.}.
\end{itemize}

We optimize the neural models for TOP1 loss using AdaGrad \cite{adagrad} with momentum for 10 epochs. Increasing the number of epochs did not significantly improve the loss in all models. We used dropout regularization \cite{srivastava14dropout} on the hidden states of \RNN{} and \HRNN{}. We applied dropout also to \GRUses{} initialization for \HRNN{} (\autoref{eq:ses_init}). We used single-layer GRU networks in both levels in the hierarchy, as using multiple layers did not improve the performance. In order to assess how the capacity of the network impacts the recommendation quality, in our experiments we considered both \textit{small} networks with 100 hidden units per \GRU{} layer and \textit{large} networks with 500 hidden units per \GRU{} layer.

We tuned the hyper-parameters of each model (baselines included) on the validation set using random search \cite{bergstra12random}. To help the reproducibility of our experiments, we report the hyper-parameters used in our experiments on the XING dataset in \autoref{tab:params} \footnote{The source code is available at \url{https://github.com/mquad/hgru4rec}.}. Dropout probabilities for \HRNN{}s are relative to the user-level \GRU{}, session-level \GRU{} and initialization in this order. The optimal neighborhood size for \ItemKNN{} is $300$ for both datasets.

Neural models were trained on Nvidia K80 GPUs equipped with 12GB of GPU memory\footnote{We employed Amazon EC2 \textit{p2.xlarge} spot instances in our experiments.}. Training times vary from $\normaltilde5$ minutes for the small \RNN{} model on XING to $\normaltilde30$ minutes for the large \HRNNAll{} on VIDEO. Evaluation took no longer than 2 minutes for all the experiments. We want to highlight that training times do not significantly differ between \RNN{} and \HRNN{}s, with \HRNNAll{} being the most computationally expensive model due to the higher complexity of its architecture (see \autoref{fig:hrnn_model}).

\begin{table}[t]
\centering
\setlength\tabcolsep{1.5pt} 
\begin{tabular}{lccP{2.5cm}}
\toprule
\textbf{Model} & \textbf{Batch size} & \textbf{Dropout} & \textbf{Learning rate/ Momentum}\\
\midrule
\RNN{} (\textit{small})$^\dagger{}$ & 100 & 0.2 & 0.1/0.1\\
\RNN{} (\textit{large})$^\dagger{}$ & 100 & 0.2 & 0.1/0.0\\
\HRNNAll{} (\textit{small}) & 100 & 0.1/0.1/0.2 & 0.1/0.2\\
\HRNNAll{} (\textit{large}) & 50 & 0.0/0.2/0.2 & 0.05/0.3\\
\HRNNInit{} (\textit{small}) & 50 & 0.0/0.1/0.0 & 0.1/0.0\\
\HRNNInit{} (\textit{large}) & 100 & 0.1/0.2/0.2 & 0.1/0.1\\
\bottomrule
\end{tabular}
\caption{Network hyper-parameters on XING ($^\dagger{}$ \RNN{} and \RNNConcat{} share the same values).}
\label{tab:params}
\end{table}

\begin{table*}[t]
\centering
\begin{tabular}{cl|ccc||ccc|}
\cline{3-8}
\multicolumn{1}{l}{} &  & \multicolumn{3}{c||}{\textbf{XING}} & \multicolumn{3}{c|}{\textbf{VIDEO}} \\ \cline{3-8} 
\multicolumn{1}{l}{} &  & \textbf{Recall@5} & \textbf{MRR@5} & \textbf{Precision@5} & \textbf{Recall@5} & \textbf{MRR@5} & \textbf{Precision@5} \\ \cline{2-8} 
\multicolumn{1}{l|}{} & \ItemKNN & 0.0697 & 0.0406 & 0.0139 & 0.4192 & 0.2916 & 0.0838 \\
\multicolumn{1}{l|}{} & \PPOP & 0.1326 & \textbf{0.0939} & 0.0265 & 0.3887 & 0.3031 & 0.0777 \\ \hline
\multicolumn{1}{|c|}{\multirow{4}{0.1cm}{\rot{\textit{small}}}} & \RNN & 0.1292 & 0.0799 & 0.0258 & 0.4639 & 0.3366 & 0.0928 \\
\multicolumn{1}{|c|}{} & \RNNConcat & \ita{0.1358} & \ita{0.0844} & \ita{0.0272} & 0.4682 & 0.3459 & 0.0936 \\
\multicolumn{1}{|c|}{} & \HRNNAll & \ita{0.1334}$^\dagger{}$ & \ita{0.0842} & \ita{0.0267}$^\dagger{}$ & \iu{0.5272} & \iu{0.3663} & \iu{0.1054} \\
\multicolumn{1}{|c|}{} & \HRNNInit & \iu{0.1337}$^\dagger{}$ & \ita{0.0832} & \iu{0.0267}$^\dagger{}$ & \iu{0.5421} & \iu{0.4119} & \iu{0.1084} \\ \hline
\multicolumn{1}{|c|}{\multirow{4}{0.1cm}{\rot{\textit{large}}}} & \RNN & 0.1317 & 0.0796 & 0.0263 & 0.5551 & 0.3886 & 0.1110 \\
\multicolumn{1}{|c|}{} & \RNNConcat & \ita{0.1467} & \ita{0.0878} & \ita{0.0293} & 0.5582 & \ita{0.4333} & \ita{0.1116} \\
\multicolumn{1}{|c|}{} & \HRNNAll & \iub{0.1482}$^\dagger{}$ & \iu{0.0925} & \iub{0.0296}$^\dagger{}$ & 0.5191 & 0.3877 & 0.1038 \\
\multicolumn{1}{|c|}{} & \HRNNInit & \ita{\textbf{0.1473}}$^\dagger{}$ & \iu{0.0901} & \iub{0.0295}$^\dagger{}$ & \iub{0.5947} & \iub{0.4433} & \iub{0.1189} \\ \hline
\end{tabular}
\caption{Results of Recall, MRR and Precision for $N=5$ on the XING and VIDEO datasets. \textit{small} networks have 100 hidden units for \RNN{}s and 100+100 for \HRNN{}s; \textit{large} networks have 500 hidden units for \RNN{}s and 500+500 for \HRNN{}s). 
All the networks have statistically significant different (ssd.) results from the baselines (Wilcoxon signed-rank test $p < 0.01$).
Networks ssd. from \RNN{} are in italic; \HRNN{}s ssd. from \RNNConcat{} are underlined. Best values are in bold. All differences between \HRNN{}s are significant excluded the values marked with $\dagger{}$ superscript.}
\label{tab:results}
\end{table*}

\subsection{Results}
We evaluate wrt. the sequential next-item prediction task, i.e. given an event of the user session, we evaluate how well the algorithm predicts the following event. All \RNN{}-based models are fed with events in the session one after the other, and we check the rank of the item selected by the user in the next event. In addition, \HRNN{} models and \RNNConcat{} are \textit{bootstrapped} with all the sessions in the user history that precede the testing one in their original order. 
This step slows down the evaluation but it is necessary to properly set the internal representations of the personalized models (e.g., the user-level representation for \HRNN{}s) before evaluation starts. Notice that the evaluation metrics are still computed only over events in the test set, so evaluation remains fair. In addition, we discarded the first prediction computed by the \RNNConcat{} baseline in each test session, since it is the only method capable of recommending the first event in the user sessions.

As recommender systems can suggest only few items at once, the relevant item should be amongst the first few items in the recommendation list. We therefore evaluate the recommendation quality in terms of Recall@5, Precision@5 and Mean Reciprocal Rank (MRR@5). In sequential next-item prediction, Recall@5 is equivalent to the hit-rate metric, and it measures the proportion of cases out of all test cases in which the relevant item is amongst the top-5 items. This is an accurate model for certain practical scenarios where no recommendation is highlighted and their absolute order does not matter, and strongly correlates with important KPIs such as CTR~\cite{itals_ecml}. Precision@5 measures the fraction of correct recommendations in the top-5 positions of each recommendation list. MRR@5 is the reciprocal rank of the relevant item, where the reciprocal rank is manually set to zero if the rank is greater than 5. MRR takes the rank of the items into account, which is important in cases where the order of recommendations matters.

\autoref{tab:results} summarizes the results for both XING and VIDEO datasets. We trained each neural model for 10 times with different random seeds and report the average results\footnote{The random seed controls the initialization of the parameters of the network, which in turn can lead substantially different results in absolute terms. Even though, we have not observed substantial differences in the relative performances of neural models when using different random seeds.}. We used Wilcoxon signed-rank test to assess significance of the difference between the proposed \HRNN{} models and the state-of-the-art session-based \RNN{} and the na\"{i}ve personalization strategy used by \RNNConcat{}.

\paragraph{Results on XING}
On this dataset, the simple personalized popularity baseline is a very competitive method, capable of outperforming the more sophisticated \ItemKNN{} baseline by large margins. As prior studies on the dataset have already shown, users' activity within and across sessions has a high degree of repetitiveness. This makes the generation of `non-trivial' personalized recommendations in this scenario very challenging \cite{abel16challenge}. 

This is further highlighted by the poor performance of session-based \RNN{} that is always significantly worse than \PPOP{} independently of the capacity of the network.
Nevertheless, personalized session-based recommendation can overcome its limitations and achieve superior performance in terms of Recall and Precision with both small and large networks. \HRNN{}s significantly outperform \RNNConcat{} in terms of Recall and Precision (up to +3\%/+1\% with small/large networks), and provide significantly better MRR with large networks (up to +5.4\% with \HRNNAll{}). Moreover, \HRNN{}s significantly outperform the strong \PPOP{} baseline of $\normaltilde 11\%$ in Recall and Precision, while obtaining comparable MRR. This is a significant result in a domain where more trivial personalization strategies are so effective.

The comparison between the two \HRNN{} variants does not highlight significant differences, excluded a small ($\normaltilde2\%$) advantage of \HRNNAll{} over \HRNNInit{} in MRR. The differences in terms of Recall and Precision are not statistically significant. Having established the superiority of \HRNN{}s over session-based recommendation and trivial concatenation, we resort to the VIDEO dataset to shed further light on the differences between the proposed personalized session-based recommendation solutions.

\paragraph{Results on VIDEO}
The experiments on this dataset exhibit drastically different results from the XING dataset. \ItemKNN{} baseline significantly outperforms \PPOP{}, and session-based \RNN{} can outperform both baselines by large margins. This is in line with past results over similar datasets \cite{hidasi16session,hidasi16feature}.

\RNNConcat{} has comparable Recall and Precision with respect to session-based \RNN{}s and interestingly significantly better MRR when large networks are used. This suggest that straight concatenation does not enhance the retrieval capabilities of the \RNN{} recommender but strengthens its ability in ranking items correctly.

However, \HRNNInit{} has significantly better performance than all baselines. It significantly outperforms all baselines and \RNN{}s (up to 6.5\% better Recall and 2.3\% MRR wrt.\ \RNNConcat{} with large networks). In other words, also in this scenario the more complex cross-session dynamics modeled by \HRNN{} provide significant advantages in the overall recommendation quality. We will investigate on the possible reasons for these results in the following sections. It is worth noting that \HRNNAll{} performs poorly in this scenario. We impute the context-enforcing policy used in this setting for the severe degradation of the recommendation quality. One possible explanation could be that the consumption of multimedia content (videos in our case) is a strongly session-based scenario, much stronger than the job search scenario represented in XING. Users may follow general community trends and have long-term interests that are captured by the user-level \GRU{}. However, the user activity within a session can be totally disconnected from her more recent sessions, and even from her general interests (for example, users having a strong general interest over extreme-sport videos may occasionally watch cartoon movie trailers). \HRNNInit{} models the user taste dynamics and lets the session-level \GRU{} free to exploit them according to the actual evolution of the user interests within session. Its greater flexibility leads to superior recommendation quality.

\subsubsection{Analysis on the user history length}
\label{sec:eval_histlen}
\begin{figure}[t]
\centering
\includegraphics[width=\linewidth]{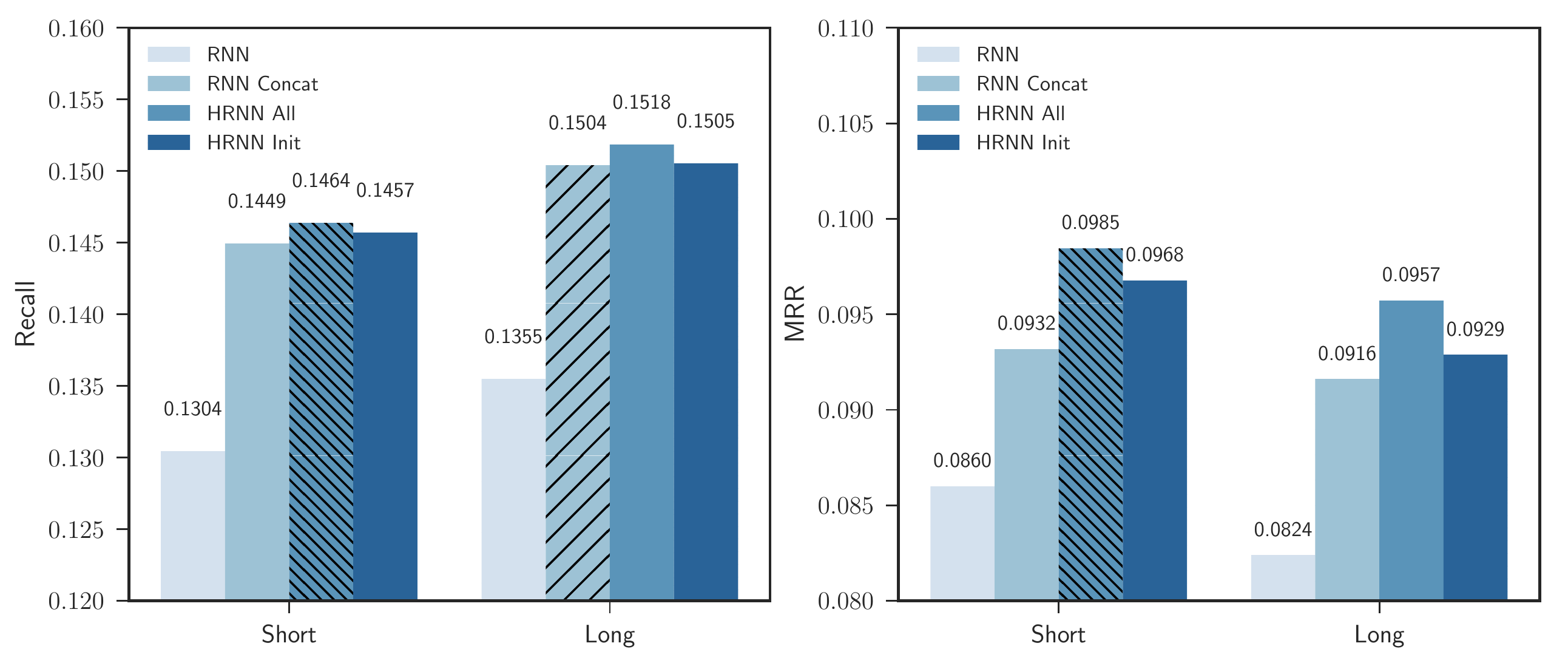}
\caption{Recall@5 and MRR@5 on XING grouped by user history length.}
\label{fig:xing_histlen}
\end{figure}

We investigate deeper the behavior of \HRNNt{} models.
Since we expect the length of the user history to have an impact on the recommendation quality, we breakdown the evaluation by the number of sessions in the history of the user. This serves as a proxy for the `freshness' of the user within the system, and allows to evaluate recommenders under different amounts of historical information about the user modeled by the user-level \GRU{} before a session begins. To this purpose, we partitioned user histories into two groups: 'Short' user histories having $\le 6$ sessions and 'Long' user histories having 6 or more. The statistics on the fraction of sessions belonging to each group for both datasets are reported in \autoref{tab:histlen_stats}. Since our goal is to measure the impact of the complex cross-session dynamics used in \HRNN{} wrt.\ traditional \RNN{}, we restrict these analysis to \RNN{}-based recommenders in the \textit{large} configuration.
For each algorithm, we compute the average Recall@5 and MRR@5 per test session grouped by history length. The analysis on Precision@5 returns similar results to Recall@5, so we omit it here also for space reasons. To enhance the robustness of the experimental results, we run the evaluation 10 times with different random seeds and report the median value per algorithm.
\begin{table}[t]
\centering
\begin{tabular}{lcc}
\toprule
\textbf{History length} & \textbf{XING} & \textbf{VIDEO} \\
\midrule
Short ($\leq 6$) & 67.00\% & 53.69\% \\
Long ($> 6$) & 33.00\% & 46.31\% \\
\bottomrule
\end{tabular}
\caption{Percentage of sessions in each history length group.}
\label{tab:histlen_stats}
\end{table}
\autoref{fig:xing_histlen} shows the results on XING. As the length of the user history grows, we can notice that Recall slightly increases and MRR slightly decreases in MRR for all methods, session-based \RNN{} included. The relative performance between methods does not changes significantly between short and long user histories, with \HRNNAll{} being the best performing model with 12\% better Recall@5 and 14-16\% better MRR@5 wrt.\ session-based \RNN{}. \HRNNInit{} has performance comparable to \RNNConcat{} and \HRNNAll{} in accordance to our previous findings.

\begin{figure}[t]
\centering
\includegraphics[width=\linewidth]{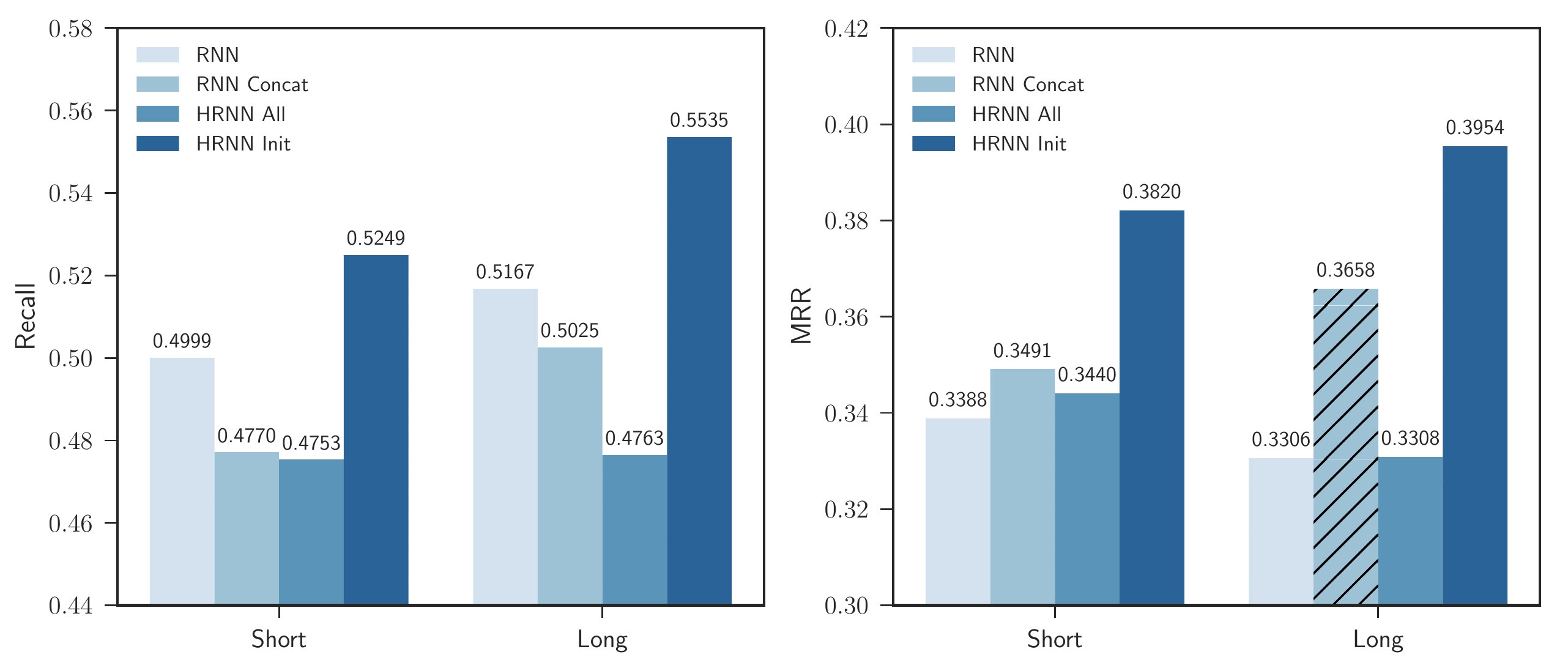}
\caption{Recall@5 and MRR@5 on VIDEO grouped by user history length.}
\label{fig:video_histlen}
\end{figure}

\autoref{fig:video_histlen} shows the results on VIDEO. Recall of all methods -- session-based \RNN{} included -- improves with the length of the user history. MRR instead improves with history length only for \RNNConcat{} and \HRNNInit{}. This highlights the need for effective personalization strategies to obtain superior recommendation quality at session-level for users that heavily utilize the system. Moreover, the performance gain of \HRNNInit{} wrt.\ session-based \RNN{} grows from 5\%/12\% (short) to 7\%/19\% (long) in Recall@5/MRR@5, further highlighting the quality of our personalization strategy. Coherently with our previous findings, \HRNNAll{} does not perform well in this scenario, and its performances are steady (or even decrease) between the two groups.

In summary, the length of the user history has a significant impact on the recommendation quality as expected. In loosely session-bounded domains like XING, in which the user activity is highly repetitive and less diverse across sessions, enforcing the user representation in input at session-level provides slightly better performance over the simpler initialization-only approach. However, in the more strongly session-based scenario, in which the user activity across sessions has a higher degree of variability and may significantly diverge from the user historical interest and tastes, the simpler and more efficient \HRNNInit{} variant has significantly better recommendation quality.

\subsubsection{Analysis within sessions}
\begin{figure}[t]
\centering
\includegraphics[width=1.02\linewidth]{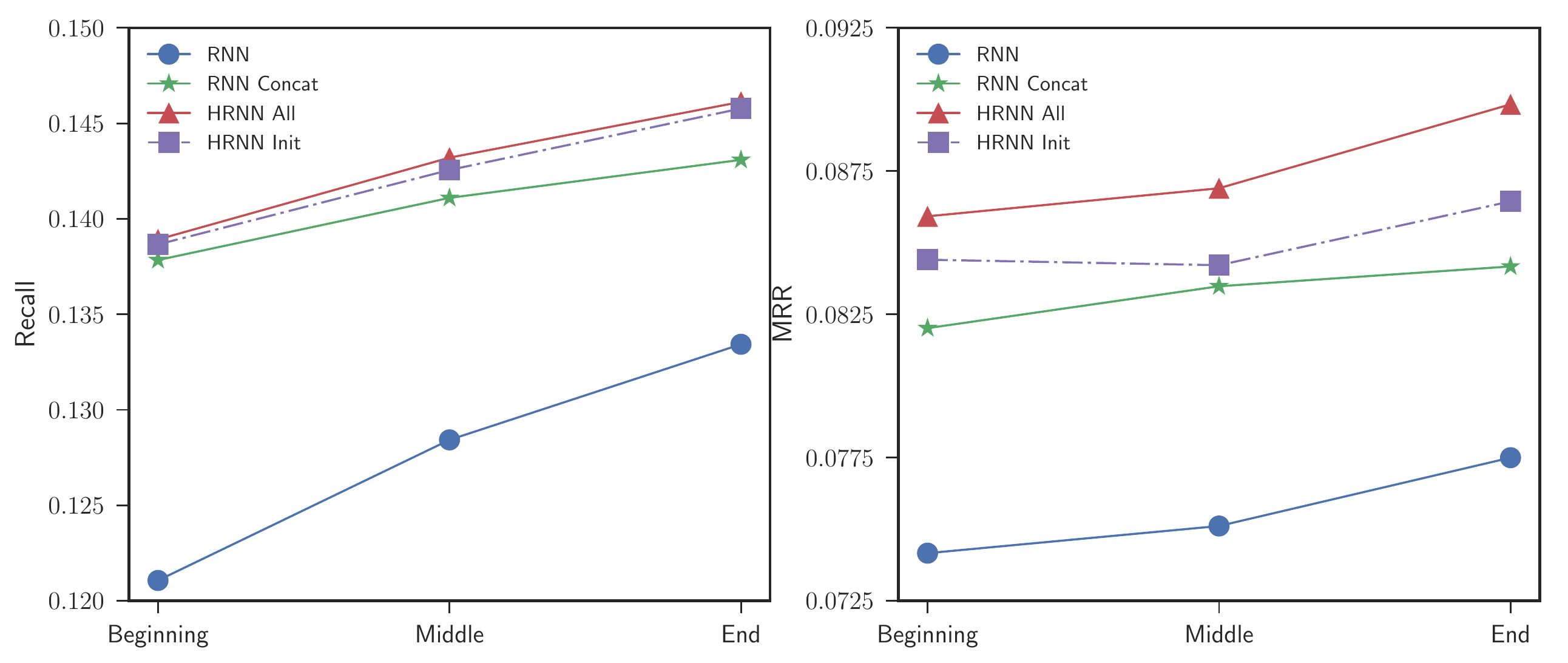}
\caption{Recall@5 and MRR@5 on XING for different positions within session.}
\label{fig:xing_sessionpos}
\end{figure}

\begin{figure}[t]
\centering
\includegraphics[width=1.02\linewidth]{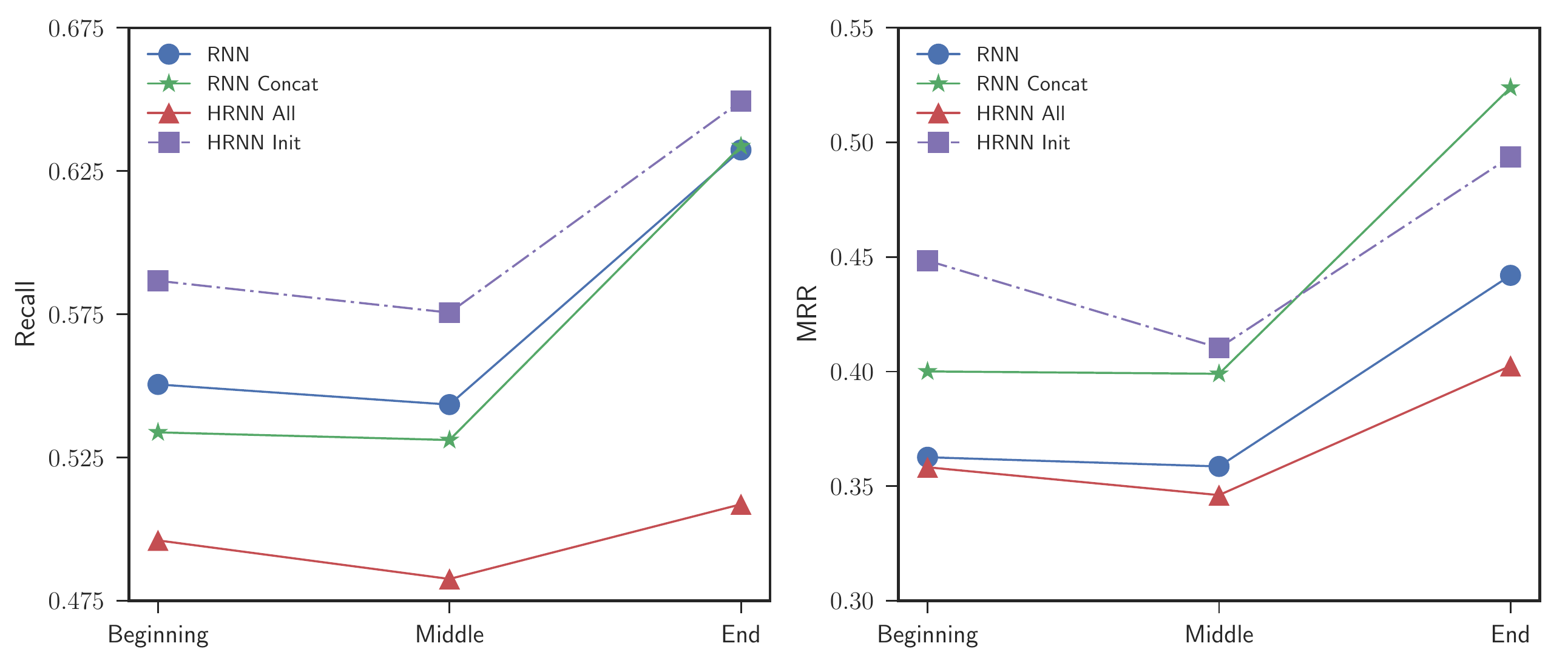}
\caption{Recall@5 and MRR@5 on VIDEO for different positions within session.}
\label{fig:video_sessionpos}
\end{figure}

Here we breakdown by number of events within the session in order to measure the impact of personalization within the user session.
We limit the analysis to sessions having length $\ge 5$ (6,736 sessions for XING and 8,254 for VIDEO). We compute the average values of each metric groped by position within the session (Beginning, Middle and End). Beginning refers to the first 2 events of the session, Middle to the 3rd and 4th, and End to any event after the 4th. As in the previous analysis, we focus on \RNN{}-based models in the \textit{large} configuration and report the median of the averages values of each metric computed over 10 runs with different random seeds. Results for Recall@5 and MRR@5 are shown in \autoref{fig:xing_sessionpos} and \autoref{fig:video_sessionpos} for XING and VIDEO respectively.

On XING, the performance of all methods increses with the number of previous items in the session, suggesting that the user context at session-level is properly leveraged by all \RNN{}-based models. However, there is a wide margin between personalized and `pure' session-based models. Both \HRNN{}s have similar Recall@5 and are comparable to \RNNConcat{}. Interestingly, the gain in MRR@5 of \HRNNAll{} wrt. to both \RNN{} and \RNNConcat{} grows with the number of items processed, meaning that in this scenario the historical user information becomes more useful as the user session continues. \HRNNInit{} has constantly better MRR that \RNNConcat{}, with wider margins at the beginning and at the and of the session.

On VIDEO, the behavior within session is different. We can notice that both Recall and MRR increase between the beginning and end of a session as expected. \HRNNInit{} exhibits a large improvement over \RNN{} and \RNNConcat{} at the beginning of the session (up to 10\% better Recall and 25\% better MRR). This conforms with the intuition that past user activity can be effectively used to predict the first actions of the user in the forthcoming session with greater accuracy. After the first few events the gain in Recall of personalized over pure session-based models reduces, while the gain in MRR stays stable. In other words, after a few events, the session-level dynamics start to prevail over longer-term user interest dynamics, making personalization strategies less effective. However, personalization still provides superior ranking quality all over the session, as testified by the higher MRR of both \HRNNInit{} and \RNNConcat{} over \RNN{}. It is important to notice that better recommendations at beginning of a session are more impactful than later in the session because they are more likely to increase the chances of retaining the user. Finally, \HRNNAll{} is always the worse method, further underpinning the superiority of the \HRNNInit{} variant.
\subsubsection{Experiments on a large-scale dataset}
We validated \HRNN{}s over a larger version of the VIDEO dataset used in the previous experiments. This dataset is composed by the interactions of 810k users on 380k videos over the same 2 months periods, for a total of 33M events and 8.5M sessions. We then applied the same preprocessing steps used for the VIDEO dataset. We named this dataset VIDEOXXL. Due to our limited computational resources, we could only test small networks (100 hidden units for \RNN{} and for 100+100 \HRNN{}s) on this large-scale dataset. We run all \RNN{}s and \HRNN{}s once using the same hyper-parameters learned on the small VIDEO dataset. Although not optimal, this approach provides a first approximation on the applicability of our solution under a more general setting. For the same reason, we do not provide an exhaustive analysis on the experimental results as done for the smaller datasets.
To speed up evaluation, we computed the rank of the relevant item compared to the 50,000 most supported items, as done in \cite{hidasi16feature}. 

Results are summarized in \autoref{tab:res_videoxxl} and confirm our previous findings on the small VIDEO dataset. \RNNConcat{} is not effective and performs similarly to session-based \RNN{}. On the other hand, \HRNNt{}s outperform session-based \RNN{} by healty margins. In particular, \HRNNInit{} outperforms session-based \RNN{} by $\normaltilde28\%$ in Recall@5 and by $\normaltilde41\%$ in MRR@5. These results further confirm the effectiveness of the \HRNN{} models presented in this paper, and further underpin the superiority of \HRNNInit{} over the alternative approaches for personalized session-based recommendation.

\begin{table}[t]
\centering
\begin{tabular}{l|c|c|c|}
\cline{2-4}
 & \multicolumn{3}{c|}{\textbf{VIDEOXXL}} \\ \cline{2-4} 
 & \textbf{Recall@5} & \textbf{MRR@5} & \textbf{Precision@5} \\ \hline
\multicolumn{1}{|l|}{\RNN} & 0.3415 & 0.2314 & 0.0683 \\
\multicolumn{1}{|l|}{\RNNConcat} & 0.3459 & 0.2368 & 0.0692 \\
\multicolumn{1}{|l|}{\HRNNAll} & 0.3621 & 0.2658 & 0.0724 \\
\multicolumn{1}{|l|}{\HRNNInit} & \textbf{0.4362} & \textbf{0.3261} & \textbf{0.0872} \\ \hline
\end{tabular}
\caption{Results on the VIDEOXXL dataset for small networks. Best values are in bold.}
\label{tab:res_videoxxl}
\end{table}

\section{Conclusions and future work}
\label{sec:conclusion}
%
In this paper we addressed the problem of personalizing session-based recommendation by proposing a model based \HRNNt{}, that extends previous \RNN{}-based session modeling with one additional \GRU{} level that models the user activity across sessions and the evolution of her interests over time. \HRNN{}s provide a seamless way of transferring the knowledge acquired on the long-term dynamics of the user interest to session-level, and hence to provide personalized session-based recommendations to returning users. The proposed \HRNN{}s model significantly outperform both state-of-the-art session-based \RNN{}s and the other basic personalization strategies for session-based recommendation on two real-world datasets having different nature. In particular, we noticed that the simpler approach that only initializes the session-level representation with the evolving representation of the user (\HRNNInit{}) gives the best results.
We delved into the dynamics of session-based \RNN{} models within and across sessions, providing extensive evidences of the superiority of the proposed \HRNNInit{} approach and setting new state-of-the-art performances for personalized session-based recommendation.

As future works, we plan to study attention models and the usage of item and user features as ways of refining user representations and improve session-based recommendation even further. We also plan to investigate personalized session-based models in other domains, such as music recommendation, e-commerce and online advertisement.


\bibliographystyle{plain}
\bibliography{paper}
\end{document}